\documentclass[10pt,twocolumn,letterpaper]{article}

\usepackage[pagenumbers]{cvpr}      %

\makeatletter
\g@addto@macro\normalsize{%
  \setlength\abovedisplayskip{8pt plus 2pt minus 5pt}%
  \setlength\belowdisplayskip{8pt plus 2pt minus 5pt}}
\makeatother

\usepackage{multirow}
\usepackage{colortbl}

\definecolor{cvprblue}{rgb}{0.21,0.49,0.74}
\usepackage[pagebackref,breaklinks,colorlinks,allcolors=cvprblue]{hyperref}

\def\confName{CVPR}
\def\confYear{2027}

\title{ShoWorld: Shape-Faithful Multi-Hypothesis Occupancy World Model}

\author{Zhiqiang Yan \quad Guangfa Wang \quad Yuan Wu \quad Jian Yang \quad Gim Hee Lee
}

\begin{document}
\maketitle
\begin{abstract}
Occupancy world models struggle to forecast moving objects due to motion uncertainty and shape distortion.
A single trajectory can average plausible futures into an implausible path, and repeated re-voxelization can erode or inflate object shape even under correct motion.
We introduce ShoWorld to address these challenges through separate motion and shape representations.
The motion forecaster predicts probabilistic trajectories conditioned on scene layout, nearby agents and the future ego trajectory.
Each hypothesis is rendered into occupancy and fused according to its probability to retain alternative futures.
The shape component predicts sub-voxel points with occupancy probabilities and offsets from local geometry.
These points move rigidly with the object, and voxel occupancy follows the probability that at least one occupied point falls within it.
This representation reduces shape distortion under motion.
ShoWorld achieves state-of-the-art results on Occ3D-nuScenes.
With ground-truth occupancy and ego trajectory, it improves mIoU and IoU by 6.56 and 9.16 points overall and by 6.06 and 5.53 points on dynamic classes.
\end{abstract}
    
\section{Introduction}
\label{sec:intro}

\begin{figure}[t]
 \centering
 \includegraphics[width=0.8\columnwidth]{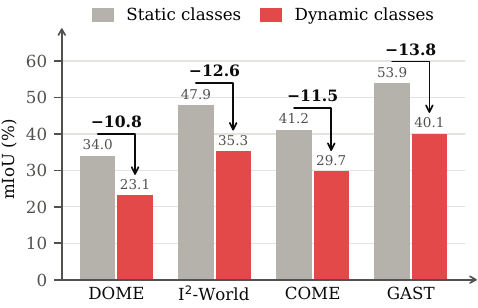}\\
 \caption{Static and dynamic mIoU of occupancy world models on Occ3D-nuScenes~\cite{tian2023occ3d}. Moving objects remain harder to forecast than static regions, and the gap widens with the prediction horizon.}
 \label{fig1}
\end{figure}

Occupancy world models forecast the evolution of 3D semantic occupancy and provide a unified representation for prediction, planning and simulation in autonomous driving~\cite{zheng2024occworld,li2025semi,yan2025drivingsphere,bian2024dynamiccity}.
Most generate future scenes in a compressed latent space through autoregressive token prediction~\cite{zheng2024occworld,jin2025occtens}, large language models~\cite{wei2026occllama,xu2025occ} or diffusion transformers~\cite{wang2024occsora,gu2024dome,li2025uniscene}.
Dedicated scene tokenizers further support this representation~\cite{yan2025renderworld,liao2025i2}.
Recent methods separate ego motion from scene changes through pose-based warping of static voxels or scene features~\cite{zhang2024efficient,chen2026geometry}, or forecast the scene in the current frame before generation~\cite{shi2026come}.
GAST~\cite{chen2026geometry} models the remaining dynamics implicitly through pose-conditioned feature modulation and attention.
These advances improve forecasting accuracy, but moving objects remain substantially harder to predict than static regions.
\cref{fig1} shows that dynamic mIoU falls 10.8--13.8 points below static mIoU.
The gap grows from 7--9 points in the first second to 11--18 points in the third.

\begin{figure*}[t]
 \centering
 \includegraphics[width=\textwidth]{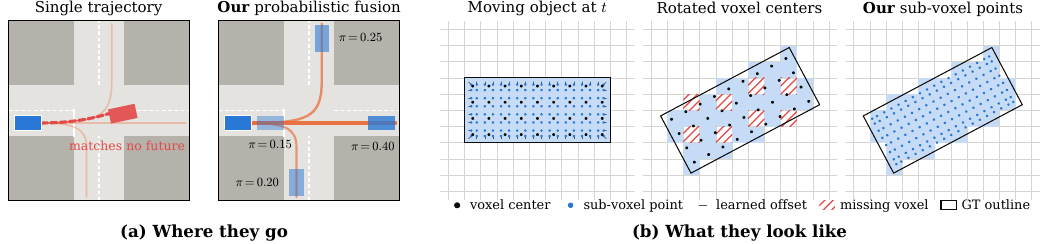}
\caption{Motion uncertainty and shape preservation.
(a) Voxel-wise supervision can pull a single trajectory between plausible futures. Our fusion retains each hypothesis with its probability.
(b) Re-voxelization of rotated voxel centers leaves holes. Sub-voxel points improve shape coverage with offsets shown as intended module outputs.} 
\vspace{-1mm}
 \label{fig:concept}
\end{figure*}

We identify two challenges behind this gap: predicting where objects move and preserving their shape under motion.
\cref{fig:concept}(a) illustrates the ambiguity of future motion.
A vehicle approaching an intersection may continue straight or turn, and a pedestrian at the curb may wait or cross.
\cref{fig:concept}(b) shows that even correct motion does not guarantee correct occupancy.
An object represented by voxels must be re-voxelized after a rotation or a translation that does not align with the grid.
This discretization can create holes or alter object extent, with errors that grow over the prediction horizon.
Our analysis shows that re-voxelization under ground-truth motion still misplaces about 22\% of moving-object voxels within the first second and 35\% in the third.

Existing methods generally couple motion and shape within scene-level prediction.
Most generate future occupancy without explicit object trajectories~\cite{zheng2024occworld,wei2026occllama,xu2025occ,yan2025renderworld,wang2024occsora,gu2024dome,li2025uniscene,jin2025occtens,liao2025i2,shi2026come,chen2026geometry}.
Others model motion through a single deterministic flow field~\cite{zhang2024efficient}.
Generative models can represent diverse futures but do not explicitly separate uncertainty in object motion from errors in object shape.
When one deterministic prediction must explain several plausible motions, voxel-wise supervision can favor a compromise between valid outcomes or spread occupancy across them.
Joint supervision of motion and shape compounds this ambiguity: a network can blur or enlarge an object to compensate for its uncertain position.
The central challenge is therefore to represent alternative motions without allowing positional uncertainty to distort the shape.

We introduce ShoWorld to address motion uncertainty and shape preservation through separate components.
The motion forecaster predicts multiple future trajectories and their probabilities for each moving object.
Each object uses the surrounding drivable space and obstacles as context.
Relative poses guide attention to nearby agents and the future trajectory of the ego vehicle.
Multimodal prediction is well established in trajectory forecasting~\cite{CoRL2020_MultiPath,CVPR2020_CoverNet,NeurIPS2022_MTR}, where hypotheses are evaluated against annotated trajectories.
ShoWorld connects these hypotheses to occupancy prediction by rendering the object under each hypothesis and fusing the results according to their probabilities.
The occupancy loss thus supervises both trajectories and probabilities.
This formulation retains the occupancy contribution of each plausible motion without forcing distinct trajectories into an intermediate path.

The shape component represents each object with sub-voxel points whose occupancy probabilities and offsets are inferred from local geometry.
These points move with the object, and the renderer computes each voxel occupancy probability from the probability that at least one occupied point falls within it.
This finer spatial coverage reduces discretization artifacts under translation and rotation.
We supervise shape under ground-truth motion to prevent trajectory errors from driving geometric distortion.
Motion hypotheses therefore represent uncertainty about future position while the shape component learns the object geometry.

Our contributions are as follows.
\begin{itemize}
    \item We propose a multi-hypothesis motion forecaster conditioned on scene layout, nearby agents and the ego trajectory. Probability-weighted fusion of renderings connects multimodal motion prediction to occupancy supervision.

    \item We identify re-voxelization as a source of shape degradation even under ground-truth motion and propose a sub-voxel point representation with separate shape supervision to reduce geometric distortion.

    \item ShoWorld sets a state of the art with gains of up to 6.06 points in dynamic mIoU and 5.53 points in dynamic IoU.
\end{itemize}

\begin{figure*}[t]
 \centering
 \includegraphics[width=\textwidth]{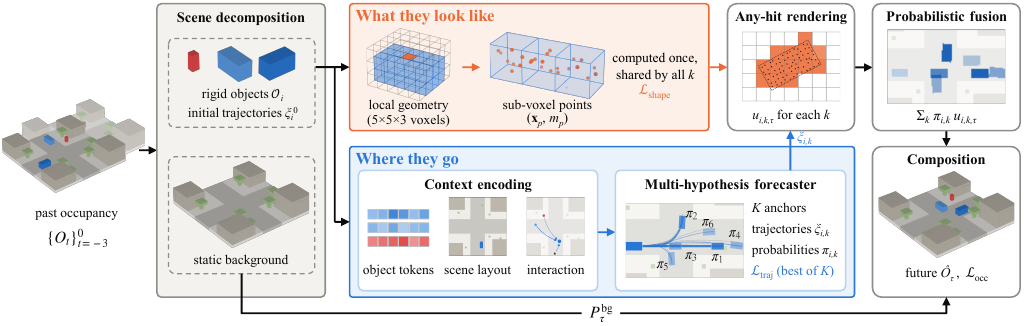}
 \vspace{-15pt}
 \caption{
 Overview of ShoWorld. The scene is decomposed into a static background and rigid objects grouped from the current occupancy. Each object predicts $K$ trajectories with probabilities from the layout, nearby agents, and the ego vehicle. Its sub-voxel points, inferred once from local geometry, are rendered under every hypothesis, fused by the probabilities, and composited over the background.
 }\label{fig2}
\end{figure*}

\section{Related Work}
\label{sec:related}

\noindent \textbf{3D Occupancy Prediction.}
3D semantic occupancy prediction estimates voxel occupancy and class labels.
The task extends semantic scene completion on SemanticKITTI~\cite{behley2019semantickitti} to surround-view scenes through dense benchmarks~\cite{tian2023occ3d,wang2023openoccupancy,wei2023surroundocc} on nuScenes~\cite{caesar2020nuscenes}.
\textit{Camera-based} methods lift image features into 3D.
MonoScene~\cite{cao2022monoscene} connects 2D and 3D UNets through feature projection, and TPVFormer~\cite{huang2023tri} uses three orthogonal planes.
VoxFormer~\cite{li2023voxformer} densifies depth-based voxel queries with a masked autoencoder.
SurroundOcc~\cite{wei2023surroundocc}, BEVFormer~\cite{li2024bevformer} and CGFormer~\cite{yu2024context} aggregate features through spatial cross-attention.
STCOcc~\cite{liao2025stcocc} jointly predicts occupancy and scene flow.
VisHall3D~\cite{lu2025vishall3d} reconstructs visible regions before hallucinating occluded geometry.
\textit{LiDAR-based} methods complete sparse point clouds through contextual shape priors~\cite{yan2021sparse}, semantic-geometric fusion~\cite{yang2021semantic,mei2023ssc} or knowledge distillation from adjacent frames~\cite{wang2024voxel}.
\textit{Camera-LiDAR fusion} combines image semantics with point-cloud geometry.
OpenOccupancy~\cite{wang2023openoccupancy} provides multimodal baselines.
Co-Occ~\cite{pan2024co} regularizes fusion through volume rendering, and OccGen~\cite{wang2024occgen} uses diffusion for occupancy prediction.
Other methods refine view transformation through attention or LiDAR depth~\cite{zhuang2024robust,duan2025sdgocc}.
These methods reconstruct the current scene, while ShoWorld forecasts its evolution.

\smallskip
\noindent \textbf{Occupancy World Models.}
World models predict scene evolution through latent features for end-to-end driving~\cite{li2024enhancing,zheng2025world4drive} or sensor data for closed-loop simulation~\cite{yang2025drivearena,yan2025drivingsphere}.
Occupancy world models forecast 3D semantic occupancy.
Most methods generate future occupancy in a compressed latent space.
OccWorld~\cite{zheng2024occworld} forecasts VQ-VAE tokens~\cite{van2017neural} with a GPT-like spatiotemporal transformer~\cite{brown2020language}.
OccSora~\cite{wang2024occsora}, DOME~\cite{gu2024dome} and DynamicCity~\cite{bian2024dynamiccity} apply diffusion transformers~\cite{peebles2023scalable} to 4D tokens, continuous latents and HexPlanes~\cite{cao2023hexplane}, respectively.
OccLLaMA~\cite{wei2026occllama} unifies occupancy, language and action in an LLM vocabulary.
Occ-LLM~\cite{xu2025occ} uses separate codebooks for movable and static voxels.
RenderWorld~\cite{yan2025renderworld} separately encodes free and occupied space with self-supervised Gaussian labels.
UniScene~\cite{li2025uniscene} generates occupancy from layouts to condition video and LiDAR synthesis.
PreWorld~\cite{li2025semi} forecasts occupancy from cameras after pretraining with 2D labels through volume rendering.
I$^2$-World~\cite{liao2025i2} improves tokenization efficiency within and across scenes.
OccTENS~\cite{jin2025occtens} uses temporal next-scale prediction to jointly model occupancy and ego motion.
Other methods separate ego motion from scene changes.
DFIT-OccWorld~\cite{zhang2024efficient} warps static voxels with ego poses and dynamic voxels with predicted flow.
COME~\cite{shi2026come} conditions diffusion on a current-frame scene forecast through ControlNet~\cite{zhang2023adding}.
GAST~\cite{chen2026geometry} warps scene features and models residual dynamics through pose-conditioned modulation and attention.
These methods improve static reconstruction but model object motion implicitly or deterministically.
ShoWorld predicts probabilistic trajectories and sub-voxel object shapes.

\section{Method}
\label{sec:method}

\textbf{Problem Formulation.} 
Given the semantic occupancy $\{O_t\}_{t=-H+1}^{0}$ of the past $H$ frames and the ego poses, an occupancy world model forecasts the next $T$ frames $\{\hat{O}_\tau\}_{\tau=1}^{T}$, each a grid of $X\times Y\times Z$ voxels labeled with one of $C$ classes including free space.
Following~\cite{gu2024dome,liao2025i2,chen2026geometry}, we use $H=4$ and $T=6$ at 2\,Hz with the future ego trajectory given or predicted.
$\mathcal{C}_d$ denotes the dynamic classes, \ie bicycle, bus, car, construction vehicle, motorcycle, pedestrian, trailer, and truck, and $\mathcal{C}_s$ the static ones.

\smallskip
\noindent \textbf{Scene Decomposition.}
Following~\cite{zhang2024efficient,chen2026geometry}, we decompose the forecast into a static background and a set of objects.
The background is transported to each future frame with the ego pose and refined by a dense decoder into class probabilities $P^{\mathrm{bg}}_\tau(v,c)$ of each voxel $v$.
Voxels of dynamic classes in the current frame, including parked vehicles, are grouped into rigid objects $\{\mathcal{O}_i\}$ with initial trajectories $\xi^0_i$ estimated from their history, as detailed in the supplementary material.
The objects are rendered into $P^{\mathrm{obj}}_\tau(v,c)$ over the dynamic classes and composited over the background as:
\begin{equation}
\begin{aligned}
&P_\tau(v,c) = \bigl(1-\alpha_\tau(v)\bigr)\,P^{\mathrm{bg}}_\tau(v,c) + P^{\mathrm{obj}}_\tau(v,c),\\
&\alpha_\tau(v) = \sum_{c\in\mathcal{C}_d} P^{\mathrm{obj}}_\tau(v,c),
\end{aligned}
\label{eq:compose}
\end{equation}
where $P^{\mathrm{obj}}_\tau(v,c)=0$ for $c\notin\mathcal{C}_d$ and $\alpha_\tau(v)\le 1$ is the probability that $v$ is occupied by an object.
\cref{fig2} shows that no learned layer follows \cref{eq:compose}.
Each moving-object error therefore traces to its predicted motion (\cref{sec:where}) or shape (\cref{sec:what}).

\subsection{Where They Go}
\label{sec:where}

\smallskip
\noindent \textbf{Object Tokens.}
We first encode the context of each object, \ie its own state, the surrounding layout, and its interactions with nearby agents and the ego vehicle.
For each object $\mathcal{O}_i$, we compute its center $\mathbf{p}_i$, its heading $\theta_i$, and its extent from the second moments of its voxels.
The heading follows the direction of motion for clearly moving objects and the principal axis otherwise.
An object token $\mathbf{z}_i$ is then encoded from the pooled voxel features, the class distribution, the extent, the speed, and the past poses in the object frame.

\smallskip
\noindent \textbf{Scene Layout.}
We build a bird's-eye view (BEV) layout map from the current occupancy to encode surface types, obstacles and cell visibility.
The map distinguishes drivable surfaces, sidewalks, terrain and other flat regions from obstacles such as buildings, vegetation and other objects.
A lightweight convolutional encoder processes the map, and its features are sampled within a heading-aligned window that spans 8\,m behind the object, 32\,m ahead and 8\,m to each side.
Adding these features to $\mathbf{z}_i$ provides each object with information about drivable space and obstacles along its likely direction of motion.

\smallskip
\noindent \textbf{Interaction with Agents and the Ego Vehicle.}
We add an ego token encoded from the past and future ego poses, and process all tokens with $L$ self-attention layers.
Each attention logit between tokens $i$ and $j$ receives a learned bias $b_{ij}=\mathrm{MLP}(\mathbf{g}_{ij})$ computed from their relative pose:
\begin{equation}
\begin{aligned}
\mathbf{g}_{ij} = \bigl[\,&R(-\theta_i)(\mathbf{p}_j-\mathbf{p}_i),\ \sin(\theta_j-\theta_i),\\
&\cos(\theta_j-\theta_i),\ \lVert\mathbf{p}_j-\mathbf{p}_i\rVert\,\bigr],
\end{aligned}
\end{equation}
where $R(\cdot)$ is a 2D rotation and the ego token is placed at the origin with zero heading.
Expressed in the frame of the attending object, the bias yields the same interaction for the same spatial configuration anywhere in the scene.

\smallskip
\noindent \textbf{Multiple Hypotheses.}
From $\mathbf{z}_i$, we predict $K$ trajectories with learnable mode embeddings $\{\mathbf{e}_k\}$.
The $k$-th hypothesis is a sequence of planar rigid poses:
\begin{equation}
\begin{aligned}
&\xi_{i,k}(\tau) = \xi^0_i(\tau) \oplus \Delta_{i,k}(\tau),\\
&\Delta_{i,k} = f_{\Delta}\bigl(\mathbf{h}_{i,k}\bigr) + \mathbf{a}_k,
\quad
\mathbf{h}_{i,k} = \mathrm{MLP}(\mathbf{z}_i+\mathbf{e}_k),
\end{aligned}
\end{equation}
where $\oplus$ adds the translation in the object frame and the yaw, and $\mathbf{a}_k$ is a fixed motion anchor.
The anchors cover keeping the current motion, braking, accelerating, turning left or right, and turning while braking, which spreads the hypotheses from the start of training, and pedestrians receive no rotation.
The probability of each hypothesis is $\pi_{i,k}=\mathrm{softmax}_k\bigl(f_{\pi}(\mathbf{h}_{i,k})\bigr)$.

\smallskip
\noindent \textbf{Probabilistic Rendering.}
Instead of selecting the most likely hypothesis, we render the object under every hypothesis and fuse the renderings by their probabilities.
Let $u_{i,k,\tau}(v)\in[0,1]$ be the occupancy of voxel $v$ when object $i$ follows hypothesis $k$ at frame $\tau$, which is defined in \cref{sec:what}.
The occupancy of the object is:
\begin{equation}
o_{i,\tau}(v) = \sum_{k=1}^{K} \pi_{i,k}\, u_{i,k,\tau}(v),
\label{eq:fusion}
\end{equation}
which is the probability that the object occupies $v$ under the predicted distribution of its future motion.
The occupancies of all objects are summed and clipped to one to form $\alpha_\tau(v)$, which is split among classes by the labels of the points.
A vehicle that may go straight, turn, or stop thus receives probability on every branch rather than in between (\cref{fig:concept}).

\begin{figure}[t]
 \centering
 \includegraphics[width=\columnwidth]{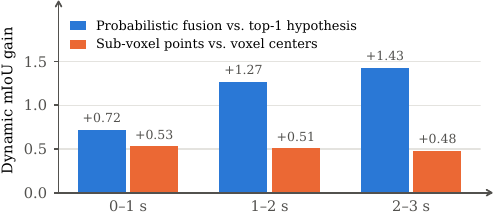}
 \vspace{-18pt}
 \caption{Gain of each component on the dynamic mIoU. Each bar replaces one component at inference, using only the most likely hypothesis or voxel centers instead of sub-voxel points.
 }\label{fig:gain}
 \vspace{-1mm}
\end{figure}

\smallskip
\noindent \textbf{Supervision.}
Since the rendering is differentiable with respect to both the poses and the probabilities, the occupancy loss supervises the trajectories and their probabilities directly, and we further supervise the hypotheses in a winner-takes-all manner.
For each object, the target pose $\xi^\ast_i(\tau)=\bigl(\mathbf{t}^\ast_i(\tau),\theta^\ast_i(\tau)\bigr)$ is the rigid transform that best aligns its voxels $\{\mathbf{x}_v\}$ with their annotated positions $\{\mathbf{x}^\ast_v(\tau)\}$,
\begin{equation}
\xi^\ast_i(\tau) = \arg\min_{\xi}\sum_{v\in\mathcal{O}_i}\bigl\lVert \xi(\mathbf{x}_v)-\mathbf{x}^\ast_v(\tau)\bigr\rVert_2^2,
\end{equation}
which has a closed-form solution.
Only voxels whose annotations agree on a single object are used, and the rotation is fixed to zero for pedestrians.
The error of hypothesis $k$ averages the translation and yaw errors over the set $\mathcal{T}_i$ of future frames in which the object is annotated,
\begin{equation}
\begin{aligned}
\ell_{i,k} = &\frac{1}{|\mathcal{T}_i|}\sum_{\tau\in\mathcal{T}_i}\Bigl(\mathrm{SL}_1\bigl(\mathbf{t}_{i,k}(\tau)-\mathbf{t}^\ast_i(\tau)\bigr)\\
&+ \lambda_{\theta}\bigl|\theta_{i,k}(\tau)\ominus\theta^\ast_i(\tau)\bigr|\Bigr),
\end{aligned}
\end{equation}
where $\mathrm{SL}_1$ is the smooth $L_1$ loss averaged over the two coordinates and $\ominus$ wraps the angle difference to $(-\pi,\pi]$.
Trajectory regression supervises only the best hypothesis $k^\ast_i=\arg\min_k \ell_{i,k}$, whose index serves as the target for probability classification:
\begin{equation}
\mathcal{L}_{\mathrm{traj}} = \frac{1}{N}\sum_{i=1}^{N}\Bigl(\ell_{i,k^\ast_i} - \lambda_{\mathrm{cls}}\log \pi_{i,k^\ast_i}\Bigr),
\end{equation}
where $N$ is the number of supervised objects.
This selective regression leaves the remaining hypotheses free to represent other plausible futures.

\begin{figure*}[t]
 \centering
 \includegraphics[width=\textwidth]{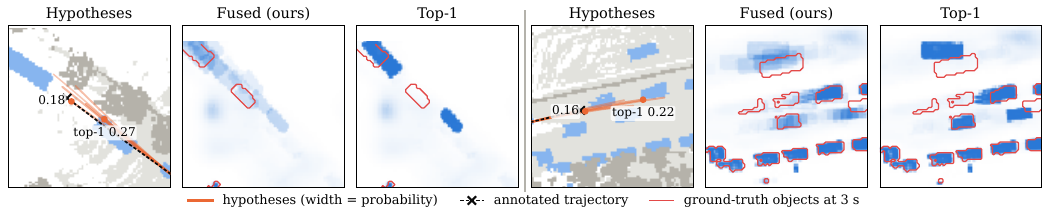}
 \vspace{-16pt}
 \caption{Two forecasts from the validation set. For each scene, we show the hypotheses of one vehicle on the current frame and the dynamic objects forecast at 3\,s by the fused hypotheses and by the most likely one. The most likely hypothesis of the vehicle ends over 4\,m from the annotated position, while another one with a probability above 0.15 ends within 2\,m.
 }\label{fig:hypotheses}
 \vspace{-1.5mm}
\end{figure*}

\smallskip
\noindent \textbf{Effect.}
At inference, fusing all hypotheses instead of rendering only the most likely one raises the dynamic mIoU by 0.72, 1.27, and 1.43 points at the three horizons (\cref{fig:gain}).
The gain grows with the horizon, where the future is the most uncertain.
\cref{fig:hypotheses} shows two cases with ambiguous vehicle speed.
The most likely hypothesis misses the annotated position at 3\,s, but the fused forecast covers it through a less likely hypothesis.

\subsection{What They Look Like}
\label{sec:what}

\noindent \textbf{Re-Voxelization.}
A common way to move an object is to transform the centers of its voxels and splat them back onto the grid.
Unless the displacement is a whole number of voxels, each voxel spreads over up to eight cells, and the object becomes thinner where the spread mass is cut by a threshold and thicker where neighboring contributions accumulate, and these errors accumulate as the object keeps moving.
The root cause is that one voxel center is too coarse a sample of the object surface.

\smallskip
\noindent \textbf{Sub-Voxel Points.}
We therefore split every voxel $v$ of an object into $s\times s$ points in the horizontal plane at regular offsets $\boldsymbol{\delta}_j$, and each point $p$ carries an occupancy $m_p\in(0,1)$ and a bounded offset:
\begin{equation}
\mathbf{x}_p = \mathbf{c}_v + \boldsymbol{\delta}_j + \rho\tanh\bigl(f_{\mathbf{x}}(\mathbf{n}_v)\bigr),
\quad
m_p = \sigma\bigl(f_{m}(\mathbf{n}_v)\bigr),
\end{equation}
where $\mathbf{c}_v$ is the voxel center, $\rho$ is half the voxel size, and the height of a point stays at its voxel center since objects move with planar rigid poses.
Both quantities are inferred from the local geometry $\mathbf{n}_v$, which labels every voxel in a $5\times5\times3$ neighborhood as the same object, another object, the static scene, or free space, together with the class of the object.
It indicates whether $v$ lies on the object boundary and identifies the boundary side.
This information guides points toward the true surface and allows points outside the object to be deactivated.
Zero initialization of the output layer places the points on a regular grid with $m_p=0.5$.
The shape is computed once in the current frame and transformed rigidly under every hypothesis to avoid cumulative re-voxelization errors.

\smallskip
\noindent \textbf{Any-Hit Rendering.}
Under hypothesis $k$ at frame $\tau$, every point is moved by the rigid pose $\xi_{i,k}(\tau)$ and by the change of ego pose, which yields $T_{i,k,\tau}(\mathbf{x}_p)$.
Let $w(\mathbf{x},v)$ be the trilinear weight of position $\mathbf{x}$ on voxel $v$.
If each point independently falls into $v$ with probability $m_p\,w(T_{i,k,\tau}(\mathbf{x}_p),v)$, a voxel is occupied with the probability that at least one point falls within it:
\begin{equation}
u_{i,k,\tau}(v) = 1-\prod_{p\in\mathcal{O}_i}\Bigl(1-m_p\,w\bigl(T_{i,k,\tau}(\mathbf{x}_p),v\bigr)\Bigr).
\label{eq:anyhit}
\end{equation}
Unlike summed splat mass, \cref{eq:anyhit} saturates as points accumulate within a voxel and remains high when a few confident points cover it.
This behavior helps preserve object extent under arbitrary motion.
The union in \cref{eq:anyhit} precedes fusion in \cref{eq:fusion} and operates within each hypothesis.
A voxel reached only by hypothesis $k$ receives probability at most $\pi_{i,k}$, regardless of how many points reach it.

\begin{figure}[t]
 \centering
 \includegraphics[width=\columnwidth]{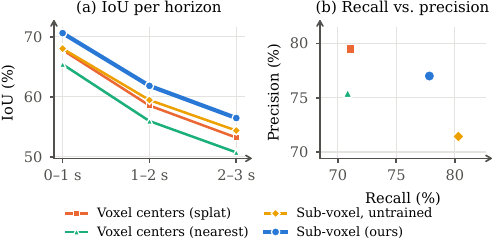}
 \vspace{-18pt}
 \caption{Shape of moving objects under ground-truth motion. Objects moved by their annotated motion are rendered by each method and compared with ground truth.
 }\label{fig:shape}
 \vspace{-3mm}
\end{figure}

\begin{table*}[t]
    \caption{4D occupancy forecasting on Occ3D-nuScenes. Methods are grouped by input modality and by whether the ego trajectory is predicted (Pred.) or given as ground truth (GT). Avg. denotes the average over the 1\,s, 2\,s, and 3\,s horizons. Results are taken from~\cite{chen2026geometry}.}
    \vspace{-8pt}
    \centering
    \footnotesize
    \renewcommand\arraystretch{0.95}
    \resizebox{\textwidth}{!}{
    \begin{tabular}{l|cc|ccc>{\columncolor[gray]{0.92}}c|ccc>{\columncolor[gray]{0.92}}c|c}
        \toprule
        \multirow{2}{*}{Method} & \multirow{2}{*}{Input} & \multirow{2}{*}{Ego traj.} &
        \multicolumn{4}{c|}{mIoU (\%) $\uparrow$} &
        \multicolumn{4}{c|}{IoU (\%) $\uparrow$} & \multirow{2}{*}{Venue} \\
        & & & 1s & 2s & 3s & Avg. & 1s & 2s & 3s & Avg. & \\
        \midrule
        OccWorld-D~\cite{zheng2024occworld} & Camera & Pred. & 11.55 & 8.10 & 6.22 & 8.62 & 18.90 & 16.26 & 14.43 & 16.53 & ECCV 2024 \\
        OccWorld-T~\cite{zheng2024occworld} & Camera & Pred. & 4.68 & 3.36 & 2.63 & 3.56 & 9.32 & 8.23 & 7.47 & 8.34 & ECCV 2024 \\
        OccWorld-S~\cite{zheng2024occworld} & Camera & Pred. & 0.28 & 0.26 & 0.24 & 0.26 & 5.05 & 5.01 & 4.95 & 5.00 & ECCV 2024 \\
        OccWorld-F~\cite{zheng2024occworld} & Camera & Pred. & 8.03 & 6.91 & 3.54 & 6.16 & 23.62 & 18.13 & 15.22 & 18.99 & ECCV 2024 \\
        RenderWorld~\cite{yan2025renderworld} & Camera & Pred. & 2.83 & 2.55 & 2.37 & 2.58 & 14.61 & 13.61 & 12.98 & 13.73 & ICRA 2025 \\
        OccLLaMA~\cite{wei2026occllama} & Camera & Pred. & 10.34 & 8.66 & 6.98 & 8.66 & 25.81 & 23.19 & 19.97 & 22.99 & ICRA 2026 \\
        PreWorld~\cite{li2025semi} & Camera & Pred. & 12.27 & 9.24 & 7.15 & 9.55 & 23.62 & 21.62 & 19.63 & 21.62 & ICLR 2025 \\
        Occ-LLM~\cite{xu2025occ} & Camera & Pred. & 11.28 & 10.21 & 9.13 & 10.21 & 27.11 & \underline{24.07} & 20.19 & 23.79 & ICRA 2025 \\
        DFIT-OccWorld~\cite{zhang2024efficient} & Camera & Pred. & 13.38 & 10.16 & 7.96 & 10.50 & 19.18 & 16.85 & 15.02 & 17.02 & arXiv 2024 \\
        GAST-STC~\cite{chen2026geometry} & Camera & Pred. & \underline{19.32} & \underline{13.96} & \underline{10.62} & \underline{14.84} & \underline{27.97} & 23.35 & \underline{20.22} & \underline{23.87} & ECCV 2026 \\
        \textbf{ShoWorld} & Camera & Pred. & \textbf{20.72} & \textbf{14.67} & \textbf{11.06} & \textbf{15.59} & \textbf{29.51} & \textbf{24.17} & \textbf{20.43} & \textbf{24.79} & - \\
        \midrule
        DOME-STC~\cite{gu2024dome} & Camera & GT & 17.79 & 14.23 & 11.58 & 14.53 & 26.39 & 23.20 & 20.42 & 23.33 & arXiv 2024 \\
        I$^2$-World-STC~\cite{liao2025i2} & Camera & GT & 21.67 & 18.78 & 16.47 & 18.97 & 30.55 & 28.76 & 26.99 & 28.77 & ICCV 2025 \\
        GAST-STC~\cite{chen2026geometry} & Camera & GT & \underline{22.90} & \underline{19.93} & \underline{17.17} & \underline{20.16} & \underline{31.62} & \underline{29.92} & \underline{28.00} & \underline{29.87} & ECCV 2026 \\
        \textbf{ShoWorld} & Camera & GT & \textbf{24.30} & \textbf{21.57} & \textbf{19.18} & \textbf{21.79} & \textbf{32.86} & \textbf{30.93} & \textbf{29.08} & \textbf{30.99} & - \\
        \midrule
        Copy\&Paste~\cite{zheng2024occworld} & 3D-Occ & Pred. & 14.91 & 10.54 & 8.52 & 11.33 & 24.47 & 19.77 & 17.31 & 20.52 & ECCV 2024 \\
        OccWorld~\cite{zheng2024occworld} & 3D-Occ & Pred. & 25.78 & 15.14 & 10.51 & 17.14 & 34.63 & 25.07 & 20.18 & 26.63 & ECCV 2024 \\
        OccLLaMA~\cite{wei2026occllama} & 3D-Occ & Pred. & 25.05 & 19.49 & 15.26 & 19.93 & 34.56 & 28.53 & 24.41 & 29.17 & ICRA 2026 \\
        RenderWorld~\cite{yan2025renderworld} & 3D-Occ & Pred. & 28.69 & 18.89 & 14.83 & 20.80 & 37.74 & 28.41 & 24.08 & 30.08 & ICRA 2025 \\
        Occ-LLM~\cite{xu2025occ} & 3D-Occ & Pred. & 24.02 & 21.65 & \underline{17.29} & 20.99 & 36.65 & 32.14 & \underline{28.77} & 32.52 & ICRA 2025 \\
        COME~\cite{shi2026come} & 3D-Occ & Pred. & 30.57 & 19.91 & 13.38 & 21.29 & 36.96 & 28.26 & 21.86 & 29.03 & NeurIPS 2025 \\
        DFIT-OccWorld~\cite{zhang2024efficient} & 3D-Occ & Pred. & 31.68 & 21.29 & 15.18 & 22.71 & 40.28 & 31.24 & 25.29 & 32.27 & arXiv 2024 \\
        GAST~\cite{chen2026geometry} & 3D-Occ & Pred. & \underline{38.60} & \textbf{24.79} & \textbf{18.54} & \underline{27.38} & \underline{44.69} & \underline{34.21} & \textbf{28.85} & \underline{35.90} & ECCV 2026 \\
        \textbf{ShoWorld} & 3D-Occ & Pred. & \textbf{42.05} & \underline{24.61} & 17.04 & \textbf{27.68} & \textbf{48.83} & \textbf{34.76} & 27.71 & \textbf{37.21} & - \\
        \midrule
        DOME~\cite{gu2024dome} & 3D-Occ & GT & 35.11 & 25.89 & 20.29 & 27.10 & 43.99 & 35.36 & 29.74 & 36.36 & arXiv 2024 \\
        UniScene~\cite{li2025uniscene} & 3D-Occ & GT & 35.37 & 29.59 & 25.08 & 31.76 & 38.34 & 32.70 & 29.09 & 34.84 & CVPR 2025 \\
        COME~\cite{shi2026come} & 3D-Occ & GT & 42.75 & 32.97 & 26.98 & 34.23 & 50.57 & 43.47 & 38.36 & 44.13 & NeurIPS 2025 \\
        I$^2$-World~\cite{liao2025i2} & 3D-Occ & GT & 47.62 & 38.58 & 32.98 & 39.73 & 54.29 & 49.43 & 45.69 & 49.80 & ICCV 2025 \\
        GAST~\cite{chen2026geometry} & 3D-Occ & GT & \underline{55.54} & \underline{46.33} & \underline{40.18} & \underline{47.40} & \underline{60.77} & \underline{55.97} & \underline{51.96} & \underline{56.24} & ECCV 2026 \\
        \textbf{ShoWorld} & 3D-Occ & GT & \textbf{63.46} & \textbf{52.66} & \textbf{45.85} & \textbf{53.96} & \textbf{70.24} & \textbf{64.88} & \textbf{61.03} & \textbf{65.40} & - \\
        \bottomrule
    \end{tabular}
    }
\label{tab:main}
\vspace{-2mm}
\end{table*}

\smallskip
\noindent \textbf{Shape Supervision with Ground-Truth Motion.}
Occupancy supervision alone can encourage the shape module to enlarge objects to compensate for trajectory errors.
Thus, we introduce shape supervision under ground-truth motion to separate geometric accuracy from trajectory accuracy.
The target poses $\xi^\ast_i(\tau)$ transform the points, which \cref{eq:anyhit} renders into $u^\ast_{\tau}(v)$.
Let $y_\tau(v)\in\{0,1\}$ indicate whether voxel $v$ belongs to a ground-truth dynamic class.
The region $\mathcal{R}_\tau$ includes all voxels reached by the transformed points and their one-voxel neighborhood.
The loss combines the binary cross-entropy and the Dice loss on $\mathcal{R}_\tau$:
\begin{equation}
\begin{aligned}
\mathcal{L}_{\mathrm{shape}} = & \frac{1}{|\mathcal{T}|}\sum_{\tau\in\mathcal{T}}\Bigl[\,\frac{1}{|\mathcal{R}_\tau|}\sum_{v\in\mathcal{R}_\tau}\mathrm{BCE}\bigl(u^\ast_\tau(v),y_\tau(v)\bigr) \\
&+ 1-\frac{2\sum_{v\in\mathcal{R}_\tau}u^\ast_\tau(v)\,y_\tau(v)}{\sum_{v\in\mathcal{R}_\tau}u^\ast_\tau(v)+\sum_{v\in\mathcal{R}_\tau}y_\tau(v)}\,\Bigr],
\end{aligned}
\end{equation}
where $\mathcal{T}$ contains the frames at 1, 2, and 3\,s.
Restricting the loss to $\mathcal{R}_\tau$ excludes unrelated objects outside the rendered neighborhood.
The fixed target poses ensure that this loss evaluates shape independently of predicted motion.

\smallskip
\noindent \textbf{Effect.}
\cref{fig:shape} isolates shape reconstruction from trajectory prediction by comparing rendering methods under ground-truth motion for annotated validation object.
Transforming voxel centers leaves gaps in the rendered shape, with IoU declining from 67.8\% within the first second to 53.2\% in the third.
Untrained sub-voxel points recover missing voxels but inflate object extent.
Our learned points retain most of this recall gain while restoring precision.
This balance improves IoU over voxel centers by 2.8, 3.3 and 3.3 points at the three horizons.
\cref{fig:gain} confirms that these shape improvements benefit the full forecast despite trajectory errors.
Replacing learned points with voxel centers at inference reduces dynamic mIoU by 0.53, 0.51 and 0.48 points at the three horizons.

\subsection{Training Objective}
\label{sec:loss}

The model is trained end-to-end with:
\begin{equation}
\mathcal{L} = \mathcal{L}_{\mathrm{occ}} + \lambda_{\mathrm{traj}}\mathcal{L}_{\mathrm{traj}} + \lambda_{\mathrm{shape}}\mathcal{L}_{\mathrm{shape}} + \mathcal{L}_{\mathrm{base}}.
\label{eq:total}
\end{equation}
Following GAST~\cite{chen2026geometry}, the occupancy loss $\mathcal{L}_{\mathrm{occ}}$ combines class-weighted cross-entropy with the Lov\'asz-softmax loss~\cite{berman2018lovasz}.
It supervises the composited occupancy $P_\tau$ over $T$ future frames and the reconstruction of $H$ past frames.
The differentiable renderer propagates this supervision to object trajectories, hypothesis probabilities and shapes.
The base loss $\mathcal{L}_{\mathrm{base}}$ combines the ego pose losses of~\cite{liao2025i2,chen2026geometry} with supervision for object grouping and initial motion.
The supplementary material provides details.
We use $K=6$, $L=2$ and $s=2$.
The loss weights are $\lambda_{\mathrm{traj}}=\lambda_{\mathrm{shape}}=\lambda_{\theta}=1$ and $\lambda_{\mathrm{cls}}=0.5$.
These weights remain fixed in training and across all experiments.
Future annotations enter only the loss computation and are excluded from the forward pass.

\begin{table*}[t]
    \caption{Long-term forecasting results on the Occ3D-nuScenes dataset over 8\,s with GT 3D occupancy and ego trajectory as inputs.}
    \vspace{-8pt}
    \centering
    \renewcommand\arraystretch{0.98}
    \setlength{\tabcolsep}{4pt}
    \resizebox{0.99\textwidth}{!}{
    \begin{tabular}{l|cccccccc>{\columncolor[gray]{0.92}}c|cccccccc>{\columncolor[gray]{0.92}}c}
        \toprule
        \multirow{2}{*}{Method} &
        \multicolumn{9}{c|}{mIoU (\%) $\uparrow$} &
        \multicolumn{9}{c}{IoU (\%) $\uparrow$} \\
        & 1s & 2s & 3s & 4s & 5s & 6s & 7s & 8s & Avg. & 1s & 2s & 3s & 4s & 5s & 6s & 7s & 8s & Avg. \\
        \midrule
        DOME~\cite{gu2024dome} & 30.10 & 21.35 & 17.36 & 14.86 & 12.61 & 11.03 & 10.00 & 9.34 & 15.83 & 39.04 & 31.20 & 27.14 & 24.73 & 22.32 & 20.28 & 19.05 & 17.97 & 25.21 \\
        COME~\cite{shi2026come} & 33.78 & 24.57 & 21.35 & 18.25 & 15.84 & 13.85 & 12.99 & 11.96 & 19.07 & 44.20 & 36.25 & 32.86 & 30.03 & 26.93 & 24.70 & 23.30 & 21.44 & 29.96 \\
        GAST~\cite{chen2026geometry} & \underline{36.23} & \underline{30.61} & \underline{27.13} & \underline{24.36} & \underline{21.98} & \underline{19.81} & \underline{17.83} & \underline{15.96} & \underline{24.12} & \underline{49.09} & \underline{45.11} & \underline{42.11} & \underline{39.52} & \underline{37.19} & \underline{34.99} & \underline{32.87} & \underline{30.87} & \underline{39.06} \\
        \textbf{ShoWorld} & \textbf{63.82} & \textbf{52.68} & \textbf{45.60} & \textbf{39.43} & \textbf{35.55} & \textbf{32.26} & \textbf{28.62} & \textbf{26.07} & \textbf{40.50} & \textbf{70.30} & \textbf{64.61} & \textbf{60.41} & \textbf{53.81} & \textbf{50.08} & \textbf{46.73} & \textbf{41.82} & \textbf{38.86} & \textbf{53.33} \\
        \bottomrule
    \end{tabular}
    }
\label{tab:long_term}
\vspace{-8pt}
\end{table*}

\begin{figure*}[t]
 \centering
 \includegraphics[width=\textwidth]{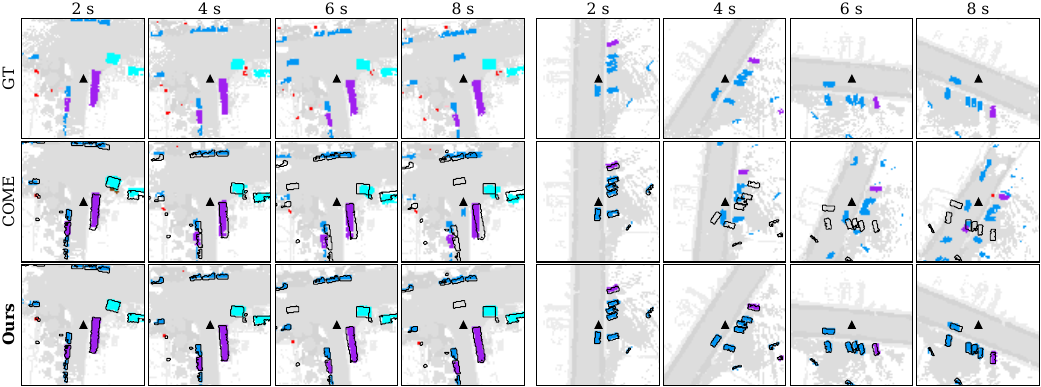}
 \vspace{-18pt}
 \caption{Qualitative comparison with COME~\cite{shi2026come} for long-term forecasting on Occ3D-nuScenes, shown for two scenes in bird's-eye view. Black outlines mark the GT positions of dynamic objects, static classes are drawn in gray, and the triangle marks the ego vehicle.
 }\label{fig:longterm_vis}
\end{figure*}

\section{Experiments}
\label{sec:experiments}

\textbf{Dataset.}
Following prior work~\cite{zheng2024occworld,gu2024dome,shi2026come,liao2025i2,chen2026geometry}, we evaluate on Occ3D-nuScenes~\cite{tian2023occ3d}, which annotates nuScenes~\cite{caesar2020nuscenes} with occupancy at 2\,Hz.
We use 700 training scenes and report results on 150 validation scenes.
The $200\times200\times16$ grid spans $[-40,40]$\,m along X and Y and $[-1,5.4]$\,m along Z.
Each $0.4$\,m voxel has one of 17 semantic labels or is free.
We forecast 3\,s ($T=6$) from 2\,s of history ($H=4$).
Four settings combine ground-truth (GT) or camera-estimated occupancy with GT or predicted ego trajectories~\cite{chen2026geometry}.
Zero-shot evaluation uses 202 Occ3D-Waymo validation scenes~\cite{tian2023occ3d,sun2020scalability}, with 39{,}987 frames at 10\,Hz and 15 semantic classes on the same grid.
We map these classes to Occ3D-nuScenes and evaluate at 10\,Hz and 2\,Hz (every 5th frame) following~\cite{liao2025i2,chen2026geometry}.
This protocol tests transfer across datasets and temporal sampling rates.
\textit{See the supplementary material for more details.}

\subsection{Comparison with the State of the Art}
\noindent \textbf{4D Occupancy Forecasting.}
\cref{tab:main} shows that ShoWorld achieves the best average mIoU and IoU in all four settings.
With 3D occupancy input and the GT ego trajectory, it surpasses the previous best method GAST~\cite{chen2026geometry} by 6.56 mIoU and 9.16 IoU with consistent gains at every horizon, and its IoU at 3\,s (61.03\%) even exceeds GAST at 1\,s (60.77\%).
Compared with DOME~\cite{gu2024dome} and COME~\cite{shi2026come}, which generate the whole scene in a latent space, the mIoU gains reach 26.86 and 19.73 points.
GAST models dynamics implicitly through pose-conditioned feature modulation and attention.
ShoWorld explicitly separates motion from shape through probabilistic trajectory hypotheses and sub-voxel points.
With the predicted ego trajectory, ShoWorld leads GAST by 0.30 mIoU and 1.31 IoU, mainly at 1\,s (3.45 and 4.14 points), as the errors of the predicted ego trajectory accumulate at longer horizons.
With camera input, ShoWorld improves mIoU and IoU over GAST-STC by 1.63 and 1.12 points with the ground-truth ego trajectory and by 0.75 and 0.92 points with predicted trajectory.
It achieves the best scores at every horizon in both settings.
These results confirm that gains persist with estimated occupancy.

\begin{figure}[t]
 \centering
 \includegraphics[width=\columnwidth]{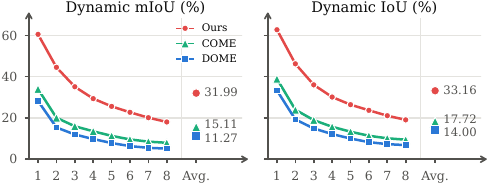}
 \vspace{-21pt}
 \caption{Long-term forecasting of dynamic classes on Occ3D-nuScenes. GAST is not included as its 8\,s weight is not released.
 }\label{fig:longterm_dynamic}
 \vspace{-9pt}
\end{figure}

\smallskip
\noindent \textbf{Long-Term Forecasting.}
Following~\cite{chen2026geometry}, we extend the forecast to 8\,s in \cref{tab:long_term}.
ShoWorld exceeds GAST by 16.38 mIoU and 14.27 IoU points on average, and its mIoU at 8\,s is even higher than that of GAST at 4\,s.
\cref{fig:longterm_dynamic} shows that the advantage holds on dynamic classes, where ShoWorld improves the average dynamic mIoU and IoU over COME by 16.88 and 15.44 points, and its values at 8\,s still exceed those of COME at 3\,s.
Our dynamic mIoU also decays more slowly, keeping 30\% of its 1\,s value at 8\,s, against 23\% and 18\% for COME and DOME.
\cref{fig:longterm_vis} illustrates both aspects.
COME gradually loses moving objects, which drift away from their GT outlines and break into fragments by 6\,s.
In contrast, our multi-hypothesis forecaster keeps each object close to its GT position up to 8\,s, even when the ego vehicle turns, showing more accurate trajectories.
The sub-voxel shape is computed once and transformed rigidly under each hypothesis.
This preserves object geometry and reduces erosion or inflation from re-voxelization.

\begin{table}[t]
    \caption{Zero-shot forecasting on Occ3D-Waymo.}
    \vspace{-8pt}
    \centering
    \renewcommand\arraystretch{0.95}
    \resizebox{0.774\columnwidth}{!}{
    \begin{tabular}{l|c|cc}
        \toprule
        Method & Rate & mIoU (\%) $\uparrow$ & IoU (\%) $\uparrow$ \\
        \midrule
        Copy\&Paste~\cite{zheng2024occworld} & 2\,Hz & 17.17 & 28.21 \\
        I$^2$-World~\cite{liao2025i2} & 2\,Hz & 36.38 & 52.36 \\
        GAST~\cite{chen2026geometry} & 2\,Hz & \underline{46.70} & \underline{60.09} \\
        \textbf{ShoWorld} & 2\,Hz & \textbf{57.84} & \textbf{68.74} \\
        \midrule
        Copy\&Paste~\cite{zheng2024occworld} & 10\,Hz & 28.34 & 40.09 \\
        I$^2$-World~\cite{liao2025i2} & 10\,Hz & 43.73 & 60.97 \\
        GAST~\cite{chen2026geometry} & 10\,Hz & \underline{57.47} & \underline{70.66} \\
        \textbf{ShoWorld} & 10\,Hz & \textbf{68.62} & \textbf{80.20} \\
        \bottomrule
    \end{tabular}
    }
\label{tab:zero_shot}
\vspace{-3pt}
\end{table}

\begin{table}[t]
    \caption{Complexity analysis on the Occ3D-nuScenes dataset.}
    \vspace{-8pt}
    \centering
    \renewcommand\arraystretch{0.95}
    \resizebox{0.95\columnwidth}{!}{
    \begin{tabular}{l|ccc|cc}
        \toprule
        \multirow{2}{*}{Method} & Params $\downarrow$ & FLOPs $\downarrow$ & Latency $\downarrow$ & mIoU $\uparrow$ & IoU $\uparrow$ \\
        & (M) & (G) & (ms) & (\%) & (\%) \\
        \midrule
        DOME~\cite{gu2024dome} & 444.07 & 2928.66 & 1899.15 & 27.10 & 36.36 \\
        COME~\cite{shi2026come} & 692.97 & 4461.46 & 4377.67 & 34.23 & 44.13 \\
        I$^2$-World~\cite{liao2025i2} & 22.67 & 494.58 & 227.09 & 39.73 & 49.80 \\
        GAST~\cite{chen2026geometry} & \underline{12.09} & \underline{416.27} & \textbf{80.03} & \underline{47.40} & \underline{56.24} \\
        \textbf{ShoWorld} & \textbf{7.93} & \textbf{123.22} & \underline{84.91} & \textbf{53.96} & \textbf{65.40} \\
        \bottomrule
    \end{tabular}
    }
\label{tab:complexity}\vspace{-5mm}
\end{table}

\smallskip
\noindent \textbf{Zero-Shot Forecasting.}
Following~\cite{liao2025i2,chen2026geometry}, we apply the model trained on Occ3D-nuScenes to Occ3D-Waymo without fine-tuning.
As shown in \cref{tab:zero_shot}, ShoWorld surpasses GAST by 11.14 mIoU and 8.65 IoU points at 2\,Hz and by 11.15 and 9.54 points at 10\,Hz.
The mIoU margins at both rates are even larger than that on Occ3D-nuScenes, suggesting that our gains generalize beyond the training dataset.
Our forecast also exceeds Copy\&Paste by about 40 mIoU points at both rates, so the model captures how the scene changes instead of repeating the current frame.
We attribute this to our explicit structure, since rigid motion, layout, relative poses, and local geometry are shared across datasets.

\smallskip
\noindent \textbf{Complexity Analysis.}
\cref{tab:complexity} shows the complexity comparison of our method with others, using a single GeForce RTX 3090. ShoWorld has only 7.93\,M parameters and 123.22\,G FLOPs, 34\% and 70\% fewer than GAST, while achieving the highest accuracy.
Its parameters are also 56 and 87 times fewer than those of DOME and COME.
This efficiency comes from the scene decomposition, in which the background is transported with the ego pose and only refined, and the dynamic part is handled by a small number of object tokens.
Its latency of 84.91\,ms is 2.7 times lower than that of I$^2$-World and comparable to that of GAST, as the object-level operations consist of many small kernels bound by launch overhead rather than arithmetic.

\subsection{Ablation Studies}
\label{sec:ablation}

All ablations in \cref{tab:ablation_main} are conducted on Occ3D-nuScenes with 3D occupancy input and the GT ego trajectory, forecasting the next 3\,s from 2\,s of history, and report the mIoU and the dynamic mIoU averaged over 1\,s, 2\,s, and 3\,s.
Baseline (a) retains the scene decomposition but omits context encoding.
Each object moves rigidly along an initial trajectory estimated from its observed history.
The transformed voxel centers are splatted back onto the grid.
Variant (b) adds context encoding to refine trajectories from object state, scene layout and interactions.
This improves mIoU and dynamic mIoU by 1.71 and 0.62 points, respectively.
These gains show the benefit of context beyond motion history.
Variants (c)--(e) increase the hypothesis count from one in (b) to 2, 4 and 6.
Dynamic mIoU improves at each tested count.
With six hypotheses, mIoU and dynamic mIoU exceed (b) by 3.49 and 6.34 points, respectively.
The dynamic mIoU gain is the largest among all components and highlights the importance of representing alternative futures.
This confirms that fusing hypotheses preserves the multimodality of future motion, which a single trajectory averages out.
(f) and (g) replace the voxel centers with sub-voxel points.
Assigning each point to its nearest voxel in (f) improves the mIoU and the dynamic mIoU by 1.55 and 0.46 points, and the any-hit rule in (g) adds another 2.08 and 1.09 points.
This shows that sub-voxel points keep moving objects shape-faithful, and the any-hit rule further prevents erosion and inflation from re-voxelization.
Training with them gains 1.55 points from (e) to (g), more than removing them only at inference (\cref{fig:gain}).
With voxel centers at inference, (g) still reaches 45.66\% dynamic mIoU, 1.05 points above (e), as its forecaster has been trained on renderings that preserve object shapes. 
Overall, the forecaster improves predictions on where objects go and the sub-voxel shape of what they look like, raising mIoU and dynamic mIoU by 8.83 and 8.51 points. 
\textit{See the supplementary material for more ablations.}

\begin{table}[t]
\centering
\caption{Ablation studies of our method on Occ3D-nuScenes.
}\label{tab:ablation_main}
\Large
\vspace{-8pt}
\renewcommand\arraystretch{1}
\resizebox{\columnwidth}{!}{
\begin{tabular}{c|c|*{3}{>{\centering\arraybackslash}p{1.8em}}|cc|cc}
\toprule
\multirow{2}{*}{Method} & Context & \multicolumn{3}{c|}{Multi-Hypothesis} & \multicolumn{2}{c|}{Sub-Voxel Shape} & \multirow{2}{*}{mIoU (\%)} & Dynamic
\\ \cline{3-7}
 & Encoding & 2 & 4 & 6 & nearest & any-hit &  & mIoU (\%)
\\ \midrule
(a) &  &  &  &  &  &  & 45.13 & 37.65 \\
(b) & \checkmark &  &  &  &  &  & 46.84 & 38.27 \\
(c) & \checkmark & \checkmark &  &  &  &  & 48.06 & 41.19 \\
(d) & \checkmark &  & \checkmark &  &  &  & 49.35 & 42.62 \\
(e) & \checkmark &  &  & \checkmark &  &  & 50.33 & 44.61 \\
(f) & \checkmark &  &  & \checkmark & \checkmark &  & 51.88 & 45.07 \\
(g) & \checkmark &  &  & \checkmark &  & \checkmark & \textbf{53.96} & \textbf{46.16} \\
\bottomrule
\end{tabular}
}
\vspace{-8pt}
\end{table}

\section{Conclusion}
\label{sec:conclusion}

We presented ShoWorld to address two challenges in occupancy forecasting: uncertainty in future object motion and shape distortion under re-voxelization.
The motion forecaster predicts multiple trajectories and their probabilities from scene layout, nearby agents and the ego trajectory.
Probabilistic fusion of rendered hypotheses retains alternative futures in the occupancy forecast.
The shape component uses sub-voxel points and an any-hit rendering rule to preserve object geometry under motion.
Supervision under ground-truth motion separates shape reconstruction from trajectory errors.
ShoWorld achieves best results on Occ3D-nuScenes with explicit trajectories and faithful shapes.

\smallskip
\noindent \textbf{Limitations.}
Our method models each object as a rigid body moving in the ground plane, and thus misses articulated motion such as pedestrian limbs.
It also relies on objects grouped from the observed occupancy, whose errors propagate into the forecast.
Since the shape is computed once in the current frame, occluded parts remain missing and objects entering later cannot be forecast.
Fusion also stays below the best hypothesis at longer horizons.
Learning the grouping jointly with the forecast and modeling newly appearing agents are left to future work.	

\par

{
    \small
    \bibliographystyle{ieeenat_fullname}
    \bibliography{main}
}

\end{document}